\documentclass{article}

\usepackage{PRIMEarxiv}

\usepackage[utf8]{inputenc} % allow utf-8 input
\usepackage[T1]{fontenc}    % use 8-bit T1 fonts
\usepackage{hyperref}       % hyperlinks
\usepackage{url}            % simple URL typesetting
\usepackage{booktabs}       % professional-quality tables
\usepackage{amsfonts}       % blackboard math symbols
\usepackage{nicefrac}       % compact symbols for 1/2, etc.
\usepackage{microtype}      % microtypography
\usepackage{lipsum}
\usepackage{fancyhdr}       % header
\usepackage{graphicx}       % graphics
\graphicspath{{media/}}     % organize your images and other figures under media/ folder
\usepackage{natbib}
\usepackage{xcolor}
\usepackage{multirow}
\usepackage{colortbl, booktabs}
\usepackage{makecell}
\usepackage{amsmath}
\title{Unleashing the True Potential of LLMs: A Feedback-Triggered Self-Correction with Long-Term Multipath Decoding
}

\author{
  Jipeng Li\textsuperscript{1}, 
  Zeyu Gao\textsuperscript{2},
  Yubin Qi\textsuperscript{1}, 
  Hande Dong\textsuperscript{1}, 
  Weijian Chen\textsuperscript{3},
  Qiang Lin\textsuperscript{1} \\
  \textsuperscript{1}Tencent \\
  \textsuperscript{2}Tsinghua University \\
  \textsuperscript{3}Institute of Dataspace, Hefei Comprehensive National Science Center
}

\begin{document}
\maketitle

\begin{abstract}
Large Language Models (LLMs) have achieved remarkable performance across diverse tasks, yet their susceptibility to generating incorrect content during inference remains a critical unsolved challenge.  While self-correction methods offer potential solutions, their effectiveness is hindered by two inherent limitations: (1) the absence of reliable guidance signals for error localization, and (2) the restricted reasoning depth imposed by conventional next-token decoding paradigms.  To address these issues, we propose Feedback-Triggered Regeneration (FTR), a novel framework that synergizes user feedback with enhanced decoding dynamics.  Specifically, FTR activates response regeneration only upon receiving negative user feedback, thereby circumventing error propagation from faulty self-assessment while preserving originally correct outputs.  Furthermore, we introduce Long-Term Multipath (LTM) decoding, which enables systematic exploration of multiple reasoning trajectories through delayed sequence evaluation, effectively overcoming the myopic decision-making characteristic of standard next-token prediction. Extensive experiments on mathematical reasoning and code generation benchmarks demonstrate that our framework achieves consistent and significant improvements over state-of-the-art prompt-based self-correction methods.
\end{abstract}

% keywords can be removed

\section{Introduction}

% Large Language Models (LLMs) have achieved remarkable performance across a variety of tasks, including text generation, question answering, and code synthesis~\cite{achiam2023gpt,touvron2023llama,guo2025deepseek}. Despite these achievements, LLMs face significant challenges, particularly the generation of incorrect information or flawed reasoning processes, which remains a widely acknowledged problem in LLM inference~\cite{yao2023llm,liu2024exploring}.
% Self-correction mechanisms have emerged as a promising approach to enhance LLM outputs through introspective reasoning, typically facilitated by carefully designed prompts~\cite{ji2023towards, madaan2024self, kim2024language, li2024confidence, chen2023iterative, huanglarge}.
% The self-correction pipeline generally consists of two phases: first, generating an initial answer using standard LLM inference; and second, prompting the LLM to assess and revise the initial output. 

Large Language Models (LLMs) have demonstrated exceptional capabilities across diverse tasks, such as text generation, question answering, and code synthesis~\cite{achiam2023gpt,touvron2023llama,guo2025deepseek}. However, a persistent challenge lies in their tendency to produce factually incorrect information or flawed logical reasoning—a critical limitation widely documented in LLM inference research~\cite{yao2023llm,liu2024exploring}.
To address this, self-correction mechanisms have emerged as a promising solution. These mechanisms are typically implemented through carefully crafted prompting strategies~\cite{ji2023towards, madaan2024self, kim2024language, li2024confidence, chen2023iterative, huanglarge}, which guide models to introspect on their initial responses.
The self-correction workflow generally follows a two-step process:
first, generating an initial answer via standard LLM inference;
and second, prompting the model to critically evaluate the initial output for errors, inconsistencies, or logical gaps, and subsequently revise it.

% However, the effectiveness of prompt-based methods remains contentious~\cite{huanglarge,li2024confidence}. As illustrated in Figure~\ref{compare}, in three typical reasoning datasets, two of the most advanced prompt-based self-correction methods proposed frequently convert correct answers into incorrect ones during the second stage of the self-correction process. Experiments show that these prompt-based methods do not lead to improvements in model performance and may even cause a decline in accuracy across the entire dataset. In this work, we identify two key challenges inherent in the prompt-based self-correction process that contribute to this phenomenon: \textbf{C1}. The prompt-based self-correction process relies on the model to assess the correctness of its previous answers, but lacks an effective guiding signal, which may lead to unnecessary self-corrections. \textbf{C2}. The self-correction process requires deeper reasoning in LLMs. However, most current methods still rely on the next-token prediction approach, which considers only one step at a time during decoding, resulting in short-term optimal outcomes. This limitation restricts the model's capacity for more profound reasoning, thereby hindering its ability to generate improved responses during self-correction.

\begin{figure}[t]
\centering
\includegraphics[width=0.6\textwidth]{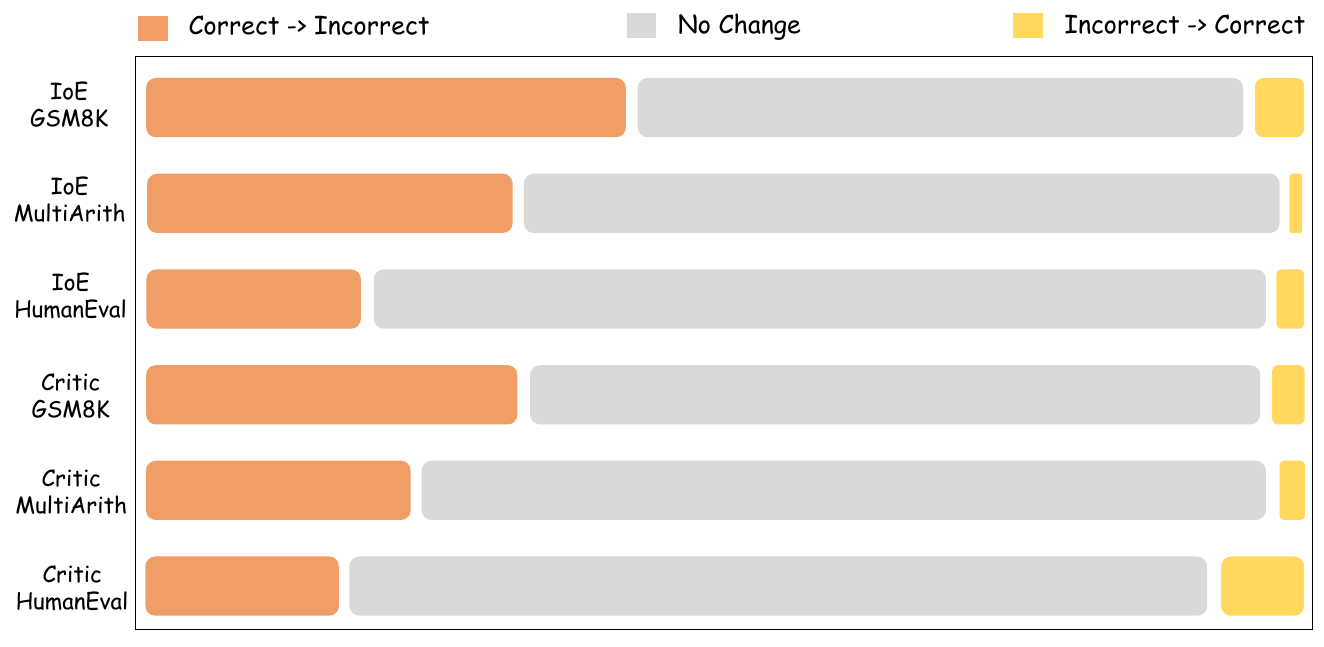} % Reduce the figure size so that it is slightly narrower than the column.
\caption{
% Percentage of answer accuracy changes resulting from prompt-based self-correction using IoE Prompts \cite{li2024confidence} and Critic Prompts \cite{huanglarge} on the Llama3-Instruct-3B model
The percentage distribution of answer changes induced by self-correction using If or Else(IoE) Prompts~\cite{li2024confidence} and Critic Prompts~\cite{huanglarge}, with experiments conducted on the Llama3-Instruct-3B.
}
\label{compare}
\end{figure}

Nevertheless, the effectiveness of prompt-based methods remains contentious.
As illustrated in Figure~\ref{compare}, experiments on three typical reasoning datasets demonstrate that two state-of-the-art prompt-based self-correction methods ~\cite{huanglarge,li2024confidence} not only frequently convert correct answers into incorrect ones but also struggle to revise incorrect answers into correct responses. In this study, we argue that this phenomenon stems from two key challenges inherent in the prompt-based self-correction process:
\begin{itemize}
\item 
\textbf{C1. Lack of Effective Guidance Signals}: The prompt-based self-correction process relies on the LLM itself to evaluate the correctness of its previous answers. However, due to the absence of explicit guidance signals, the LLM may fail to accurately determine which parts require revision, leading to unnecessary self-corrections (i.e., modifying correct answers to incorrect ones). 
% This issue is particularly pronounced in complex problems, as the LLM may be unable to distinguish between its own correct information and the erroneous information that needs correction.
Additionally, given the sensitivity of LLMs to input, biased prompts may cause incorrect alignment and mislead the LLM into making inaccurate judgments~\cite{huanglarge}.
\item 
\textbf{C2. Shallow Decoding Limits Deep Reasoning}: Self-correction of erroneous results by LLMs requires deeper thinking and reasoning. However, most current methods follow the next-token prediction paradigm, which focuses only on single-step predictions during the decoding process. The correctness of the answer depends on a comprehensive evaluation of the entire output answer sequence. This short-term, step-oriented decoding process limits the LLM's ability to engage in deeper reasoning, thereby hindering its capacity to generate improved responses during the self-correction process.

\end{itemize}

To address \textbf{C1}, we propose Feedback-Triggered Regeneration (FTR), a self-correction framework that leverages user feedback (e.g., explicit thumbs-up/down signals) to guide LLMs' reasoning refinement. In human-AI interactions, users naturally provide feedback—such as dissatisfaction cues—when engaging with LLM outputs, which acts as a critical trigger for identifying responses requiring deeper revision. By treating feedback as a revision trigger, FTR avoids unnecessary reprocessing of outputs that have received positive user signals, focusing computational resources on cases requiring improvement.
In the FTR framework, negative feedback guides the LLM to regenerate responses by reprocessing the original input, with feedback solely determining whether regeneration is initiated. This design circumvents limitations of prompt-based methods—such as self-critical prompting that relies on flawed introspection or labor-intensive prompt tuning for each task—thereby enhancing both correction efficiency and generalizability.

% To address \textbf{C2}, we propose a novel Long Vision Multipath (LTM) decoding strategy designed to promote deeper reasoning in LLMs. This method explores multiple paths at each decoding step while evaluating long-term performance, thereby expanding the search space available to LLMs. Compared to the next-token prediction approach, our method is more likely to identify sequences with higher long-term quality. To enable the model to refine its previous responses more effectively and produce more accurate and coherent outputs, this method is applied during the regeneration stage of self-correction.

% To address \textbf{C2}, we strengthen FTR by integrating a novel Long-Term Multipath (LTM) decoding strategy, which is designed to promote deeper reasoning in LLMs. Specifically, LTM explores multiple decoding paths at each step and evaluates their long-term performance, thereby overcoming the short-term optimization of standard approach that focus solely on next token probability at the expense of overall sequence quality. By considering multiple paths and their long-term impact, LTM enables the LLM to recompose its responses more effectively, leading to more accurate and coherent outputs. This strategy is applied during the regeneration stage of self-correction, allowing the LLM to generate improved responses by leveraging deeper reasoning.

To address \textbf{C2}, we enhance FTR by integrating a Long-Term Multipath (LTM) decoding strategy, which enables LLMs to perform global reasoning through adaptively multi-step path exploration. Unlike standard autoregressive decoding methods (e.g., greedy decoding) that optimize solely for next-token probability, LTM evaluates multiple decoding paths at each step, prioritizing long-term sequence coherence and semantic consistency. This approach mitigates the limitation of short-term optimization—such as local optimum traps and context drift—that plague traditional decoding frameworks. Critically, LTM is dynamically activated during the regeneration stage—triggered by negative user feedback—to replace the standard single-path decoding, thereby enabling deep reasoning for error correction without incurring excessive computational overhead.

% Overall, we propose a novel user feedback-triggered self-correction method that integrates feedback with the LLM decoding strategy to enhance outputs from both prompt and decoding perspectives. To assess the effectiveness of our framework, we conduct a series of experiments comparing it with existing prompt-based methods using open-source LLMs. These experiments focus on challenging and representative mathematical and coding datasets. Our results demonstrate that prompt-based self-correction methods often deteriorate model performance, consistent with prior findings~\citep{huanglarge}. In contrast, our method achieves over a 10\% improvement across most tasks.

% In summary, our main contributions are as follows: 
% \begin{enumerate}
%     \item We propose a novel self-correction framework that incorporates user feedback as regeneration signal.
%     \item We propose a novel decoding method that evaluates the long-term performance of multiple reasoning paths to enhance the accuracy of the generated responses. 
%     \item We conduct experiments on various datasets and models, demonstrating the superiority of our method.
% \end{enumerate}

% Overall, we propose a novel user feedback-triggered self-correction framework that integrates user feedback with the advanced LTM decoding strategy to enhance outputs from both encoding and decoding perspectives. 
Overall, we propose a novel user feedback-triggered self-correction framework that integrates user feedback with the advanced LTM decoding strategy.
To validate the effectiveness of our framework, we conduct a series of experiments comparing it with SOTA prompt-based methods using open-source backend LLMs. These experiments focused on challenging mathematical and coding datasets, where our framework consistently achieves superior performance.
In summary, our main contributions are as follows:
\begin{enumerate}
\item We introduce a novel self-correction framework that leverages user feedback as a regeneration signal, thereby preventing unnecessary self-corrections and improving the quality of LLM outputs.
\item We propose a novel decoding method that evaluates the long-term performance of multiple reasoning paths, thereby enhancing the accuracy and coherence of generated responses.
\item We demonstrate the superiority of our method through extensive experiments on various datasets and backend LLMs.
% , showing consistent improvements over existing prompt-based approaches.
\end{enumerate}

\section{Preliminaries}
% In this section, we introduce the notation used throughout this paper and provide an overview of the commonly employed two-stage self-correction framework.

\subsection{Notation Definition}
% Let $x = (x_0,x_1,...,x_n)$ denote an input sequence, and $y = (y_0,y_1,...,y_m)$ represent the corresponding LLM response, where $y = \mathcal{M}(x)$. Here $\mathcal{M}$ denotes a typical autoregressive language model. In this context, the response is generated sequentially during the decoding process. At the $i$-th step of the inference process of the LLM, we define the probability of the current output sequence $s_i$ as $P(s_i)$, which is calculated as the product of the likelihoods of the first $i$ tokens.

% \begin{equation}
%     P(s_i)=\prod_{k=1}^i P(y_k|y_{0:k-1},x)P(y_0|x),
% \end{equation}
% The perplexity (PPL) value of the sequence at step $i$ is then defined as:
% \begin{equation}
%     PPL_i = P(s_{i})^{-\frac{1}{i}}
% \end{equation}
% In this work, PPL is employed to assess the quality of a sequence during the decoding process.

Let \( x = (x_0, x_1, \ldots, x_n) \) denote an input sequence, and \( y = (y_0, y_1, \ldots, y_m) \) represent the corresponding LLM response, where \( y = \mathcal{M}(x) \). Here, \( \mathcal{M} \) denotes a typical autoregressive language model. In this context, the response is generated sequentially during the decoding process. At the \( i \)-th step of the inference process, we define the probability of the current output sequence \( s_i = (y_0, y_1, \ldots, y_i) \) as \( P(s_i) \), which is calculated as the product of the likelihoods of the first \( i \) tokens:
\begin{equation}
P(s_i) = P(y_0 | x) \prod_{k=1}^i P(y_k | y_{0:k-1}, x).
\end{equation}
The perplexity (PPL) value of the sequence at step \( i \) is then defined as:
\begin{equation}
\text{PPL}_i = P(s_i)^{-\frac{1}{i+1}},
\end{equation}
which  quantifies how confidently a language model predicts the sequence by measuring the inverse geometric mean of token probabilities, where lower values indicate better predictions. In this work, PPL is employed as a metric to assess the quality of a sequence during the decoding process~\cite{jelinek1977perplexity,brown2020language}.

\subsection{Two-Stage Framework for Self-Correction}
\begin{figure}[t]
\centering
\includegraphics[width=0.6\textwidth]{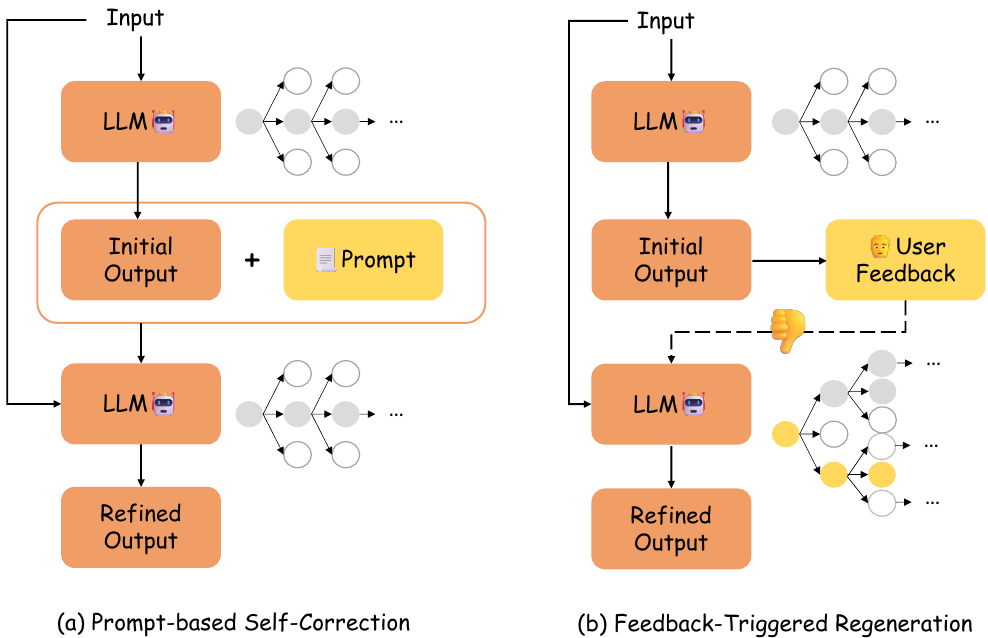} % Reduce the figure size so that it is slightly narrower than the column.
\caption{(a) Framework of the prompt-based self-correction approach.
(b) Framework of our feedback-triggered self-correction approach.}
\label{method}
\end{figure}

The framework of most prompt-based self-correction methods can be divided into two stages, as depicted in Figure \ref{method} (a): 
\begin{itemize}
    \item  \textbf{Stage 1}: An initial input $x$ is provided to the LLM to generate an initial response $y_{init} = \mathcal{M}(x)$. 
    
    \item \textbf{Stage 2}: An independent correction prompt $p_{cor}$ is then given, prompting the LLM to reflect on its generated response. This enables the LLM to refine its answer, regenerating the refined output $ y_{cor} = \mathcal{M}(x,y_{init},p_{cor})$.
\end{itemize}
% In this work, we adopt and enhance this two-stage framework by incorporating user feedback, which serves as a guiding signal for prompting the LLM to regenerate responses based on the initial prompt $x$. In the second stage, we employ LTM decoding, which utilizes PPL to select the most plausible sequences at each decoding step, ensuring the model generates more refined and accurate responses. The entire workflow is illustrated in Figure \ref{method} (b).
In Figure~\ref{method} (b), we also present our proposed FTR self-correction framework for intuitive comparison. Specifically, we have made improvements from two perspectives: 1) incorporating user feedback as a guiding signal to prompt LLM to regenerate responses based on the initial input when necessary; 2) adopting LTM decoding to enhance the LLM's ability for deeper reasoning in order to address more complex error response scenarios.

\begin{figure}[t]
\centering
\includegraphics[width=0.8\textwidth]{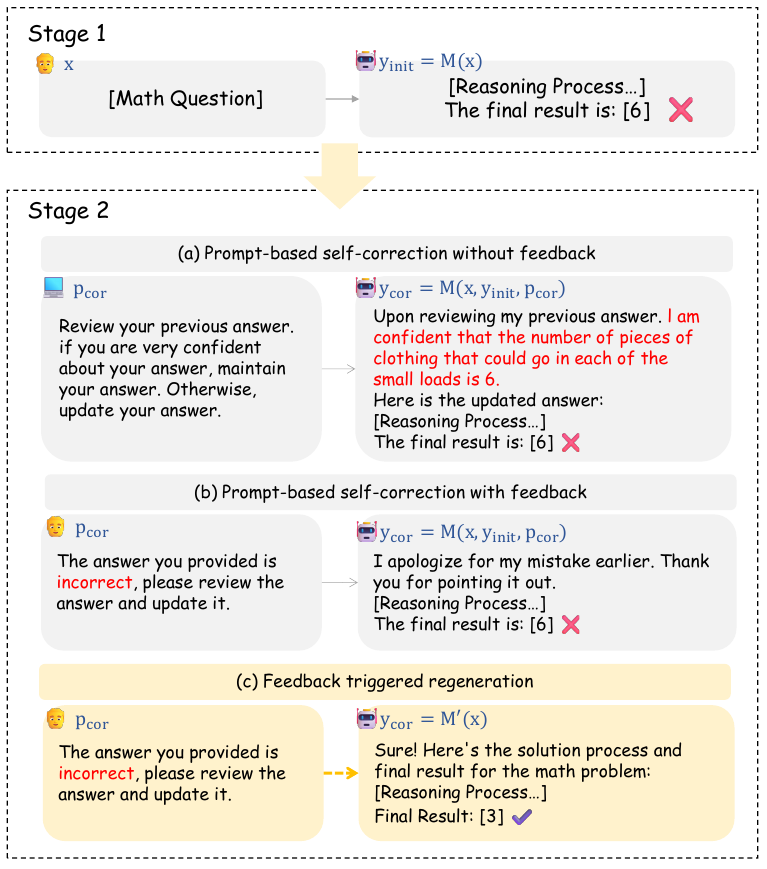} % Reduce the figure size so that it is slightly narrower than the column.
\caption{Comparison of different self-correction methods. (a) Self-assessment and update; (b) Revision with user feedback prompt; (c) Regeneration triggered by user feedback.}
\label{selfc}
\end{figure}

\section{Methodology}
% In this section, we first introduce our proposed feedback-triggered regeneration framework, followed by a detailed exploration of LTM decoding.

\subsection{Feedback-Triggered Regeneration}

% In most studies, the correction prompt $p_{cor}$ does not provide feedback on the correctness of $y_{init}$.  A key limitation of this approach, as highlighted in previous works \citep{huanglarge, madaan2024self} and illustrated in Figure \ref{compare}, is that LLMs are often unable to independently assess the correctness of their responses. Consequently, they may make erroneous decisions, as demonstrated in Figure \ref{selfc} (1). Furthermore, when the model is well-aligned and receives a carefully crafted initial input, the first response should ideally be optimal, given the selected decoding algorithm. Introducing an additional prompt may, however, cause the model to generate a response that is more aligned with the combined input, as noted by \citet{huanglarge}. Another challenge is the model’s sensitivity to prompt design \citep{xu2024take, liu2024self}, which can ultimately degrade performance. As shown in Figure \ref{selfc} (2), even when correctness feedback is included in the $p_{cor}$, the model may fail to refine the previous answer accurately.

In general, the correction prompt $p_{cor}$  provides no information about the correctness of the initial response $y_{init}$. Additionally, LLMs often lack the ability to independently assess the correctness of their own responses, as highlighted in previous works~\citep{madaan2024self,huanglarge}. These limitations may lead to erroneous decisions by the LLM, as demonstrated in Figure~\ref{selfc} (a).

While properly trained LLMs generally produce optimal initial responses to well-formed queries, our investigation identifies two key limitations in self-correction frameworks: (1) Prompt-induced solution drift, where externally introduced guidance inadvertently distorts the model's reasoning trajectory, causing systematic deviations from the original optimal solution~\citep{huanglarge}; and (2) Correction capability constraints, as evidenced in Figure~\ref{selfc}(b) showing persistent refinement failures even when explicit correctness feedback ($p_{cor}$) is provided. This dual challenge of prompt fragility and inherent correction limitations fundamentally restricts the practical effectiveness of current self-correction paradigms ~\citep{xu2024take, liu2024self}.

% To mitigate the negative impact of self-correction prompts on LLM output, we propose a FTR framework that employs the structured two-step methodology:
% \begin{itemize}
% \item \textbf{Stage 1}: An initial input $x$ is provided to guide the LLM in generating an initial response $y_{init} = \mathcal{M}(x)$. 
% \item \textbf{Stage 2}:  If user feedback indicates that the LLM's output $y_{init}$ is problematic, the original prompt $x$ and an advanced decoding strategy, LTM, are employed to regenerate the output. Otherwise, no further action is taken.
%     \begin{equation}
%         y_{cor} = \mathcal{M'}(x)
%     \end{equation}
% \end{itemize}
% Here, $ \mathcal{M'}$ denotes the model with an alternative decoding method. In our approach, human feedback serves as an indicator for triggering regeneration, rather than as an additional prompt, as shown in Figure \ref{method} (b) and Figure \ref{selfc} (c). This prevents the model output from being degraded by potentially biased prompts.
% % We use $\mathcal{M'}$ to indicate that the sampling method in the second step may differ from that in the first step.

To avoid the negative impact of self-correction prompts, we propose an enhanced two-stage FTR self-correction framework:
\begin{itemize}
    \item \textbf{Stage 1}: 
Similarly, provide the initial input $x$ to the LLM to generate the initial response $y_{init} = \mathcal{M}(x)$. 

    \item \textbf{Stage 2}:  If user feedback indicates that the LLM's output $y_{init}$ is problematic, the original prompt $x$ and an advanced decoding strategy, LTM, are employed to regenerate the output. Otherwise, no further action is taken.
    \begin{equation}
        y_{cor} = \mathcal{M'}(x)
    \end{equation}
\end{itemize}
Here, $ \mathcal{M'}$ denotes the LLM equipped with the LTM decoding strategy, which will be described in the following section. Note that the second stage of FTR uses only the original input $x$ without introducing additional prompts. Human feedback serves solely as an indicator to trigger regeneration, as illustrated in Figure~\ref{method} (b) and Figure~\ref{selfc} (c). This approach prevents the LLM output from being degraded by potentially biased prompts.

\subsection{Long-Term Multipath Decoding}

\begin{figure}[t]
\centering
\includegraphics[width=0.8\textwidth]{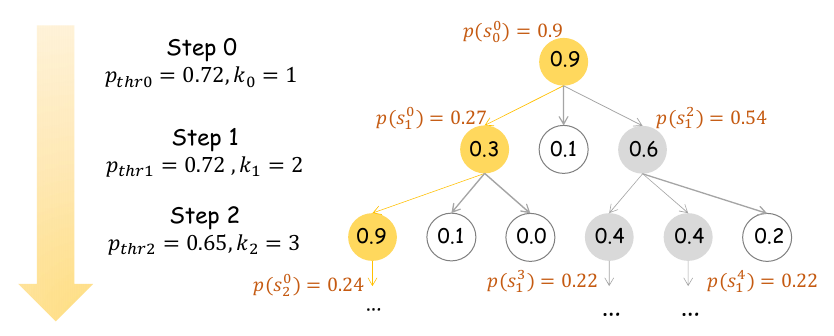} % Reduce the figure size so that it is slightly narrower than the column.
\caption{Illustration of LTM decoding strategies ($V=3$), where black numbers in circles are token likelihood and red ones indicate sequence likelihood.}
\label{LTM}
\end{figure}

When users provide negative feedback on LLM responses, it indicates the need for deeper reasoning beyond standard next-token prediction, as this conventional paradigm lacks comprehensive evaluation of complete answer quality. Therefore, we enhance the decoding process through two key mechanisms:
(1) \textbf{Multipath Exploration}: Instead of exploring a single path, token selection is performed using a ``tree'' structure rather than the traditional ``chain'' structure. This allows the LLM to explore multiple potential sequences simultaneously, as illustrated in Figure~\ref{LTM}.
(2) \textbf{Sequence Evaluation}: Using PPL as our quality metric, we dynamically retain the top \( k_i \) sequences at each step \( i \), with \( k_i \) adjusted according to the step's PPL distribution. Unlike Beam Search's constant beam width, our method dynamically adjusts the candidate count \( k_i \) during decoding.

LTM's core mechanism dynamically determines $ k_i $ through token distributions, with mathematical formalization provided below and demonstration in Figure~\ref{LTM}.
\textbf{Firstly}, the probabilities of all possible sequences are computed. Let \( k_{i-1} \) denote the number of candidates retained at the \((i-1)\)-th step, and let \( V \) represent the size of the LLM's vocabulary. Accordingly, there are \( k_{i-1} \times V \) candidate sequences.
\textbf{Secondly}, the top-\( k_i \) sequences are selected from the \( k_{i-1} \times V \) candidates. These candidates are sorted based on their probabilities \( P(s_i^j) \), where \( j \in [0, k_{i-1} \times V - 1] \) denotes the index of each candidate. The cumulative probability is computed until it exceeds the threshold value $p_{thr_i}$:
\begin{equation}
    \sum_{j=0}^{k_i} P(s_{i}^j)\geq p_{thr_i}.
\end{equation}
The number of retained sequences, denoted as $k_i$, is the minimal set that satisfies this condition , with all other sequences pruned. 
When all candidate sequences have equal length, PPL can be computed directly from the sequence probability. For computational convenience, we use $P(s_{i}^j)$ as our evaluation metric. The threshold $p_{thr_i}$ is defined as:
\begin{equation}
    p_{thr_i} = p^{*} \times \sum_{j=0}^{k_{i-1}\times V-1} P(s_{i}^j).
\end{equation}
Here, $p^{*} \in [0,1]$ controls the number of sequences to be pruned, where a lower $p^*$ results in more sequences being discarded.
\textbf{Finally}, to control computational overhead, an additional hyperparameter \( k^* \) is introduced. When \( k_i \) exceeds \( k^* \), only the first \( k^* \) sequences are retained. During decoding, LTM selects the most probable candidates subject to dual constraints \( p_{\text{thr}_i} \) and \( k^* \), ultimately selecting the optimal answer through minimum PPL evaluation from multiple generated candidates.

In traditional sampling methods that select one token at a time, errors introduced early in the process can propagate through subsequent stages. Moreover, a token initially selected as the optimal choice may lose its advantage as the context evolves, leading to suboptimal sequence selection and a decline in overall performance. 
For instance, as illustrated in Figure~\ref{LTM}, the left branch at step 1 has a lower probability than the right branch but achieves better performance at step 2. This highlights the limitations of methods that focus solely on immediate token probabilities without considering long-term sequence quality.
In contrast, LTM explores multiple sequences and evaluates the overall long-term performance of each. This approach enables the LLM to look ahead at future tokens and retrospectively correct errors, thereby enhancing its ability to generate higher quality outputs.

\section{Experimental Setup}

% In this section, we detail the experimental setup of our study, encompassing the LLMs employed, datasets utilized, baseline methods compared, evaluation metrics applied, and implementation details.

\subsection{Datasets}
To evaluate the effectiveness of our method, we select mathematical and coding datasets that require model reasoning capabilities while allowing objective verification of outputs. The detailed dataset specifications are as follows:
(1) \textbf{GSM8K} \citep{cobbe2021gsm8k}: This dataset comprises 1,319 grade-school mathematics problems with standard solutions, utilizing the zero-shot prompt: "\textit{The following is a math problem from elementary school. Please provide the solution process and the final result. Write the final result within square brackets. Only one pair of square brackets should be used, and it should contain only a number. The question is:...}"
(2) \textbf{MultiArith} \citep{roy2015solving}: This dataset contains 180 mathematical problems evaluated using identical prompting methodology to GSM8K.
(3) \textbf{HumanEval} \citep{chen2021evaluating}: This dataset includes 164 programming questions. Prompts are formulated based on implementations from the DeCLaRe Lab \citep{chia2023instructeval}.

\subsection{Baselines}

% Our framework is a two-stage process that operates independently of external tools and is applicable to a wide range of tasks. 
To evaluate the effectiveness of our method, we compare it with two general-purpose, two-stage prompt-based self-correction approaches: (1) \textbf{Critic Prompt} \citep{huanglarge}: To ensure experimental fairness, we adopt the experimental setting from \cite{li2024confidence}, which instructs the LLM to identify errors in its previous responses and generate refined results.
    % The original approach utilized a three-stage prompting mechanism: the first stage provides the initial prompt, the second stage directs the model to identify errors, and the third stage prompts the model to revise its response. 
(2) \textbf{If or Else (IoE) Prompt} \citep{li2024confidence}: 
This method prompts the LLM to assess its confidence in the initial answer and generate a refined response if necessary.

For the Critic Prompt method, the LLM is instructed using the following prompt: "\textit{Review your previous answer and find problems with your answer. Based on the problems you found, improve your answer. Please reiterate your answer}". For the IoE Prompt method, the prompt employed is "\textit{Review your previous answer. If you are very confident about your answer, maintain your answer. Otherwise, update your answer}". 
The key difference between these methods aligns with variations in $p_{cor}$ within the discussed framework.

\subsection{Evaluation metrics}
We evaluate model performance using top-1 accuracy ($acc@1$) for mathematical tasks and top-1 pass rate ($pass@1$) for coding tasks. $acc@1$ measures the percentage of test cases where the model’s highest-ranked prediction matches the ground truth, while $pass@1$ quantifies the proportion of instances where the top prediction successfully passes a predefined coding test. % The corresponding results are summarized in Table \ref{selfcorrect}.

\subsection{Implementation Details}

% To verify the universality of our method, we test various open-source LLMs ranging from 1B to 13B parameters. These include Llama2-Chat-7B and Llama2-Chat-13B \citep{touvron2023llama}, Llama3-Instruct-1B and Llama3-Instruct-3B \citep{dubey2024llama}, and Qwen2.5-1.5B-Instruct and Qwen2.5-3B-Instruct \citep{qwen2.5}. For brevity, we refer to these LLMs as Llama2-7B, Llama2-13B, Llama3-1B, Llama3-3B, Qwen-1.5B, and Qwen-3B in the subsequent sections.

% All methods employ nucleus sampling \citep{holtzman2019curious} with parameters $p=0.95, k=15$ during initial stage. While baseline approaches reuse this strategy in the second stage, our FTR framework implements LTM decoding instead. 

To verify the universality of our method, we test various open-source LLMs ranging from 1B to 13B parameters, including Llama2-Chat-7B and Llama2-Chat-13B \citep{touvron2023llama}, Llama3-Instruct-1B and Llama3-Instruct-3B \citep{dubey2024llama}, and Qwen2.5-1.5B-Instruct and Qwen2.5-3B-Instruct \citep{qwen2.5}. For brevity, we refer to these LLMs as Llama2-7B, Llama2-13B, Llama3-1B, Llama3-3B, Qwen-1.5B, and Qwen-3B in the subsequent sections. 

All methods employ nucleus sampling \citep{holtzman2019curious} with hyperparameters $p=0.95, k=15$ during initial stage. In contrast, baseline approaches retain this strategy in the second stage, whereas our FTR framework employs the proposed LTM decoding strategy—featuring multipath exploration and long-term context evaluation—during the regeneration phase.

\section{Experimental Results}

% In this section, we present the empirical findings that demonstrate the efficacy of our approach.

\subsection{Overall Comparison}

\begin{table*}[t]
  \centering
    \fontsize{9}{9}\selectfont
    \setlength{\tabcolsep}{9pt}
    \begin{tabular}{c|ccc|ccc}
    \toprule
     & \multicolumn{3}{c|}{\textbf{PROTOCOL 1}} & \multicolumn{3}{c}{\textbf{PROTOCOL 2}}\\
    Method & \textbf{GSM8K} & \textbf{MultiArith} & \textbf{HumanEval} & \textbf{GSM8K} & \textbf{MultiArith} & \textbf{HumanEval} \\
    \midrule
    \rowcolor{gray!20}
    \multicolumn{7}{c}{\# Llama2-7B \#} \\
    Initial Input & 0.206 & 0.539 & 0.104  & 0.206 & 0.539 & 0.104 \\
    + Critic Prompt & 0.171 & 0.522 & 0.043 & 0.202 & 0.572 & 0.104\\
    + IoE Prompt& 0.136 & 0.339 & 0.091  & 0.189 & 0.406 & 0.116 \\
    + FTR (Ours)& \textbf{0.360} & \textbf{0.878} & \textbf{0.165} & \textbf{0.269} & \textbf{0.744} & \textbf{0.140} \\
    \rowcolor{gray!20}
    \multicolumn{7}{c}{\# Llama2-13B \#} \\
    Initial Input & 0.303 & 0.656 & 0.207  & 0.303 & 0.656 & 0.207\\
    + Critic Prompt & 0.122 & 0.322 & 0.049 & 0.276 & 0.611 & 0.207 \\
    + IoE Prompt& 0.281 & 0.656 & 0.122 & 0.308 & 0.700 & 0.195 \\
    + FTR (Ours)& \textbf{0.463} & \textbf{0.917} & \textbf{0.250} & \textbf{0.419} & \textbf{0.822} & \textbf{0.226} \\
    % DeepSeek-1.5B& 0.569 & 0.850 & -  \\
    % DeepSeek-7B&  & & -  \\
      \rowcolor{gray!20}
    \multicolumn{7}{c}{\# Llama3-1B \#} \\
    Initial Input & 0.245 & 0.406 & 0.287 & 0.245 & 0.406 & 0.287 \\
    + Critic Prompt & 0.167 & 0.289 & 0.165 & 0.246 & 0.439 & 0.287 \\
    + IoE Prompt& 0.173 & 0.383 & 0.281  & 0.244 & 0.439 & 0.317 \\
    + FTR (Ours)& \textbf{0.399} & \textbf{0.622} & \textbf{0.409}  &\textbf{0.310} & \textbf{0.473} & \textbf{0.384} \\
    % DeepSeek-1.5B& 0.604 & 0.883 & -  \\
    % DeepSeek-7B&  & & -  \\
          \rowcolor{gray!20}
    \multicolumn{7}{c}{\# Llama3-3B\#} \\
    Initial Input & 0.774 & 0.961 & 0.488 & 0.774 & 0.961 & 0.488 \\
    + Critic Prompt & 0.485 & 0.750 & 0.390 &0.753 &0.972 &0.506 \\
    + IoE Prompt& 0.394 & 0.650 & 0.323  &0.744 &0.878 &0.482  \\
    + FTR (Ours)& \textbf{0.875} & \textbf{0.994} & \textbf{0.604}  &\textbf{0.837} &\textbf{0.983} & \textbf{0.585} \\
              \rowcolor{gray!20}
    \multicolumn{7}{c}{\# Qwen-1.5B\#} \\
        Initial Input & 0.422 & 0.678 & 0.409 & 0.422 & 0.678 & 0.409 \\
    + Critic Prompt & 0.334 & 0.506 & 0.220  &0.446 & 0.667&0.390  \\
    + IoE Prompt& 0.328 & 0.567 & 0.085  & 0.434 & 0.656 & 0.323 \\
    + FTR (Ours)& \textbf{0.594} & \textbf{0.839} & \textbf{0.732}  & \textbf{0.459} & \textbf{0.689} & \textbf{0.585} \\
              \rowcolor{gray!20}
    \multicolumn{7}{c}{\# Qwen-3B\#} \\
    Initial Input & 0.837 & 0.994 & 0.677 & 0.837 & 0.994 & 0.677 \\
    + Critic Prompt & 0.789 & 0.939 & 0.524  & 0.843 & 0.994 & 0.720 \\
    + IoE Prompt& 0.823 & 0.989 & 0.628  & 0.846 & 0.994 & 0.687 \\
    + FTR (Ours)& \textbf{0.902} & \textbf{1.000} & \textbf{0.768}  & \textbf{0.879} &\textbf{1.000} & \textbf{0.750} \\
    % DeepSeek-1.5B& \textbf{0.824} & \textbf{1.000} & -  \\
    % DeepSeek-7B&  & & -  \\
    \bottomrule
    \end{tabular}%
    \caption{Overall performance comparison of self-correction methods across different datasets and LLMs.}
  \label{selfcorrect}%
\end{table*}%

% To rigorously evaluate the performance of FTR against existing self-correction methods, we employ two complementary evaluation strategies. \textbf{First}, we compare FTR with standard prompt-based baselines under their default configurations: the baseline methods perform self-assessment and correction without external feedback, while in FTR we simulate feedback by comparing LLM initial outputs against ground-truth labels, regenerating responses only for initially incorrect answers. \textbf{Second}, to establish a fair comparison under realistic human evaluation conditions, we implement a unified evaluation framework where GPT-4o\cite{achiam2023gpt} assesses the first-stage outputs, serving as an automated proxy for human judgment. This approach ensures methodological consistency by regenerating only those responses that GPT-4o classifies as incorrect, applied equally to both FTR and baseline methods. We justify the use of GPT-4o as an evaluator based on its demonstrated competence: it achieves 76.6\% accuracy on the challenging MATH dataset, comparable to mathematically skilled humans\cite{DBLP:conf/nips/HendrycksBKABTS21}, and matches human-level performance on coding tasks\cite{DBLP:journals/corr/abs-2503-15885}. Quantitative analysis of GPT-4o's evaluation errors (false positives and negatives) is provided in Table \ref{falsenegative}. 

Given the challenge of directly obtaining real user feedback in controlled experimental settings—where subjective human preferences and contextual interactions are difficult to replicate—we design two complementary evaluation protocols to rigorously assess FTR against state-of-the-art self-correction methods:
(1) \textbf{Supervised Ground-Truth Evaluation (Protocol 1)}:
In a supervised setting, we compare FTR with prompt-based baselines that rely solely on internal self-assessment (i.e., no user or external signal). Here, baselines perform self-correction via default prompt configurations, while FTR simulates supervised feedback by leveraging ground-truth labels to identify initially incorrect answers and trigger regeneration. This protocol isolates the impact of objective error signals, ideal for tasks with verifiable ground truths (e.g., math problem solving).
(2) \textbf{Human-Mimicking Proxy Evaluation (Protocol 2)}:
To bridge the gap between controlled experiments and real-world human-AI interactions, we employ GPT-4o \cite{hurst2024gpt} as an automated proxy for human judgment. This unified framework applies GPT-4o’s error classification (e.g., factual errors, logical flaws) to first-stage outputs, mimicking how users might subjectively evaluate responses. Critically, regeneration is triggered only for outputs deemed incorrect by the proxy, ensuring fair comparison across methods. 
\footnote{
The rationality of this approach is supported by GPT-4o’s demonstrated human-like competence: 76.6\% accuracy on the MATH dataset \cite{DBLP:conf/nips/HendrycksBKABTS21} and comparable code generation performance to humans \cite{DBLP:journals/corr/abs-2503-15885}.
Quantitative analysis of the proxy’s evaluation errors (false positives/negatives) is provided in Appendices Table \ref{falsenegative}.}

Table \ref{selfcorrect} presents a comprehensive performance comparison between FTR and baseline approaches across datasets, LLMs, and both evaluation protocols. Under Protocol 1, FTR achieves substantial performance gains (10\%–20\%) consistently across all evaluated scenarios, highlighting its effectiveness in integrating user feedback with the LTM decoding strategy to drive targeted corrections. This consistency demonstrates FTR’s adaptability to diverse model scales and task types, from factual answer to logical reasoning.
In contrast, quantitative results show that baseline methods like Critic and IoE prompt techniques often exhibit performance degradation compared to their own initial outputs across most datasets and LLMs. These findings align with prior research by \cite{huanglarge}, which attributes such issues to prompts potentially misleading LLMs into altering correct responses during self-assessment. Together, these results underscore the limitations of prompt-based self-correction in open-source models, where introspective capabilities remain constrained. Under Protocol 2, FTR demonstrates consistent superiority over baselines across model architectures and task domains, even when using noisy feedback proxies. This robustness validates FTR’s effectiveness in realistic settings and confirms that improvements stem from its architectural design—rather than reliance on ground-truth labels—as shown by comparable gains in both protocols. Such cross-protocol analysis further reveals that explicit accuracy feedback (Protocol 1) and simulated human feedback (Protocol 2) both contribute to FTR’s performance, albeit through different mechanisms, reinforcing its generalizability as a self-correction framework.

% \begin{figure}[t]
% \centering
% \includegraphics[width=0.5\textwidth]{latex/scatter.png} % Reduce the figure size so that it is slightly narrower than the column.
% \caption{ Performance comparison of different self-correction methods across different datasets and LLMs with imperfect feedback.
% }
% \label{imperfect}
% \end{figure}

\subsection{Feedback Experiment}

\begin{figure}[t]
\centering
\includegraphics[width=0.8\textwidth]{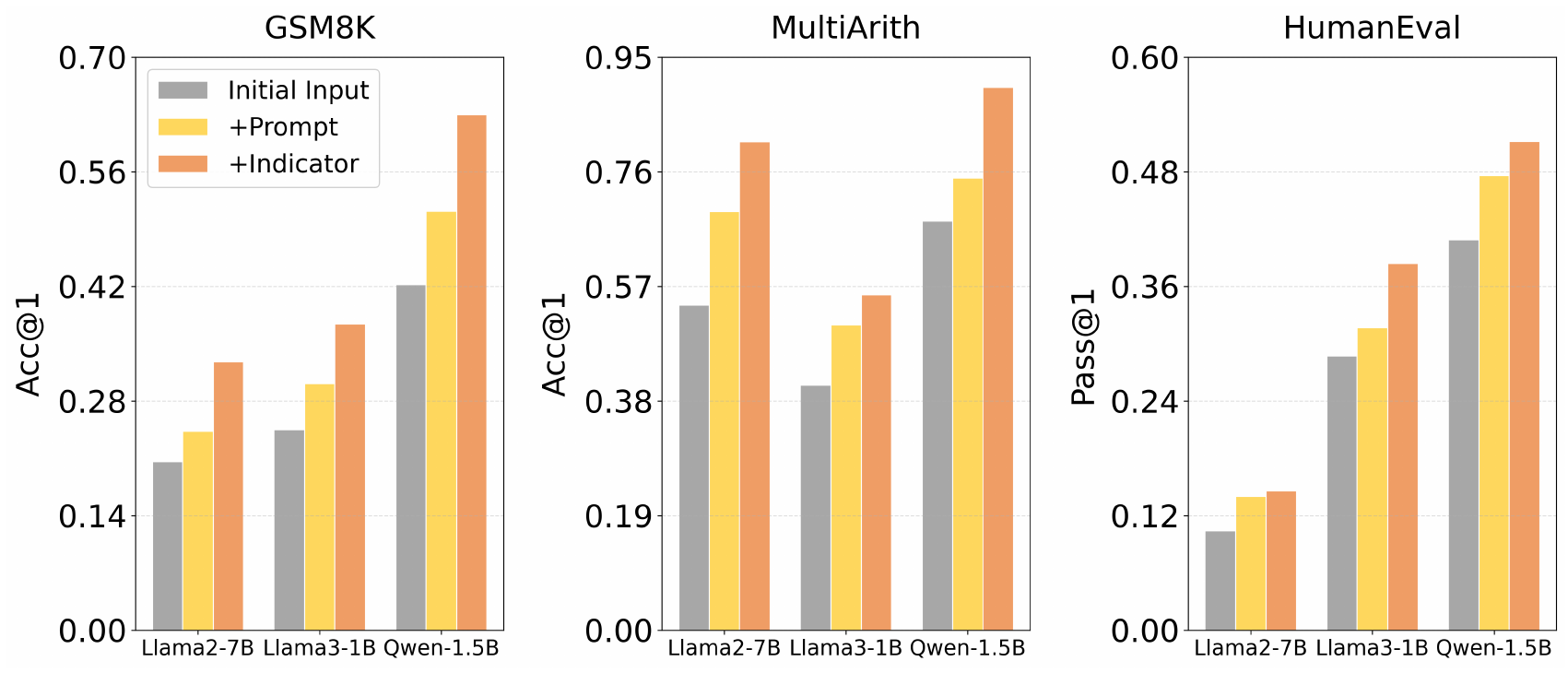} % Reduce the figure size so that it is slightly narrower than the column.
\caption{Performance comparison of different user feedback utilization approaches.}
\label{feedback_figure}
\end{figure}

Here, we analyze the design rationale of the proposed FTR framework, which leverages user feedback as a guiding signal to refine LLM outputs. A critical limitation of prompt-based self-correction methods is their propensity to erroneously revise correct responses during self-assessment—a phenomenon often attributed to flawed introspective mechanisms \cite{huanglarge}. User feedback addresses this by serving as an external validation signal, directly identifying outputs requiring revision without relying on the LLM’s internal judgment.
To investigate the optimal integration of feedback into self-correction workflows, we design two experimental configurations:
(1) \textbf{feedback-as-prompt}: The accuracy of the initial answer is encoded as a natural language prompt to guide refinement. For example, when an initial response is incorrect, the LLM receives the prompt: "\textit{The answer you provided contains factual errors. Please review the question and regenerate a correct response.}"
(2) \textbf{feedback-as-indicator}: The system uses feedback solely as a binary trigger (regenerate if feedback is negative) and reprocesses the original input without additional prompts. Both configurations employ identical nucleus sampling strategy to ensure comparability.

Figure \ref{feedback_figure} presents the comparative performance of three representative LLMs across different reasoning benchmarks. The feedback-as-indicator approach significantly outperforms feedback-as-prompt methods in reasoning accuracy, providing empirical evidence that conventional prompt-based correction mechanisms introduce decision boundary instability and lead to performance degradation.
These results empirically support the recommendation to employ user feedback primarily as a regeneration trigger—rather than embedding it in corrective prompts—to maintain decoding stability and optimize LLM self-correction efficiency.
\footnote{
Complete quantitative results for all backend LLMs and datasets are documented in Appendices Table \ref{feedbackuse}.}
\subsection{Decoding Comparison}

% \begin{figure}[t]
% \centering
% \includegraphics[width=0.50\textwidth]{latex/decoding_ablation.png} % Reduce the figure size so that it is slightly narrower than the column.
% \caption{Comparison of decoding methods across different LLM architectures on MultiArith.}
% \label{decoding_figure}
% \end{figure}

\begin{table}[t]
\centering
\begin{tabular}{cccc}
\toprule
   Decoding Strategy &  \textbf{GSM8K} & \textbf{MultiArith} & \textbf{HumanEval}  \\
\midrule
\rowcolor{gray!20}
    \multicolumn{4}{c}{\# Llama2-7B \#} \\
    Greedy Decoding & 0.243 & 0.683& 0.122\\
    Beam Search & 0.261& 0.739 & 0.134 \\
 Combined Sampling & 0.253&0.733 &0.134 \\
 Adaptive Decoding & 0.257&0.722 & 0.128 \\
 LTM Decoding & \textbf{0.276} & \textbf{0.750}& \textbf{0.146} \\
\rowcolor{gray!20}
    \multicolumn{4}{c}{\# Llama2-13B \#} \\
    Greedy Decoding &0.330 & 0.750& 0.220\\
    Beam Search & 0.366& 0.800& 0.220 \\
 Combined Sampling&0.345 & 0.756&0.213 \\
  Adaptive Decoding & 0.366& 0.722& 0.213 \\
  LTM Decoding & \textbf{0.378} & \textbf{0.833}& \textbf{0.232} \\
\rowcolor{gray!20}
    \multicolumn{4}{c}{\# Llama3-1B \#} \\
    Greedy Decoding & 0.286 & 0.428 & 0.274 \\
    Beam Search & 0.268& 0.478 & \textbf{0.354} \\
 Combined Sampling & 0.231&0.411 &0.287 \\
  Adaptive Decoding &0.257 &0.322 &0.348  \\
  LTM Decoding & \textbf{0.289} & \textbf{0.494}& \textbf{0.354} \\
\rowcolor{gray!20}
    \multicolumn{4}{c}{\# Llama3-3B \#} \\
    Greedy Decoding &0.782 &0.967 &0.500 \\
    Beam Search & 0.796& 0.983 & 0.555 \\
 Combined Sampling &0.761 & 0.950&0.506 \\
  Adaptive Decoding & 0.747&0.972 &0.512  \\
  LTM Decoding & \textbf{0.804} & \textbf{0.994}& \textbf{0.561} \\
\rowcolor{gray!20}
    \multicolumn{4}{c}{\# Qwen-1.5B \#} \\
    Greedy Decoding &0.441 & 0.456&0.457 \\
    Beam Search &\textbf{0.456} &0.489  & 0.610  \\
 Combined Sampling &0.418 &0.439 &0.390 \\
  Adaptive Decoding & 0.439&0.506 &0.476  \\
  LTM Decoding & \textbf{ 0.456} & \textbf{0.522 }& \textbf{0.622} \\
\rowcolor{gray!20}
    \multicolumn{4}{c}{\# Qwen-3B \#} \\
    Greedy Decoding &0.842 &1.000 & 0.756\\
    Beam Search &0.851 &\textbf{1.000}  & 0.756 \\
Combined Sampling &0.817 &0.994 &0.762 \\
  Adaptive Decoding &0.823 &0.994 &0.713  \\
  LTM Decoding & \textbf{0.852} & \textbf{1.000}& \textbf{0.768} \\
\bottomrule
\end{tabular}
\caption{Performance comparison of different decoding strategies for backend LLMs.}
\label{ablation}
\end{table}

To evaluate the LTM method, we assess its performance as a standalone decoding strategy for LLMs without integrating any self-correction techniques, focusing on single-turn tasks. The models and datasets used are consistent with those described in the preceding sections.
The baseline methods comprise: (1) \textbf{Greedy Decoding}: This method always selects the word with the highest probability at each decoding step. (2) \textbf{Beam Search}: This method selects the top-$k$ most probable beams at each decoding step and is a classical decoding technique in the field of NLP.
(3) \textbf{Combined Sampling (nucleus sampling)}: This approach combines top-$p$ and top-$k$ sampling, with parameters $p=0.95$ and $k=15$ used consistently across all experiments. It is a widely adopted decoding strategy in LLMs.
(4) \textbf{Adaptive Decoding}: This method enhances top-$k$ sampling by dynamically adjusting the candidate set size at each generation step based on an entropy-based confidence score. Adaptive Decoding is a relatively novel technique compared to the aforementioned methods.

% To ensure a fair comparison, we equalize the computational budgets (measured by the number of generated tokens) by adjusting the hyperparameters of each method (see Appendix Table \ref{hyperparameter}). The results are shown in Table \ref{ablation}.  It is evident that LTM outperforms all baseline methods. This highlights the efficacy of multi-path exploration and dynamically adjusted candidate sets, which can extend the exploration of the decoding space. Compared to Beam Search, LTM can also ensure that computational resources are concentrated on the most crucial steps. For example, in the case of a flattened token distribution, LTM can select more tokens compared to those in highly concentrated distributions, thereby achieving better overall performance than Beam Search's fixed token strategy. 

To ensure fair comparative evaluation, we standardize computational budgets—quantified by generated token counts—across methods through hyperparameter adjustments.\footnote{
Detailed hyperparameter settings for different decoding  strategies are presented in Appendices Table \ref{hyperparameter}.
}
Results in Table \ref{ablation} demonstrate that LTM consistently outperforms baseline approaches. This superiority highlights the effectiveness of its multipath exploration and dynamically adjusted candidate set mechanism, which expand the decoding space exploration beyond single-path methods. Compared to Beam Search, LTM strategically allocates computational resources to critical decoding steps. For example, in scenarios with flattened token probability distributions, LTM adaptively selects diverse tokens, whereas Beam Search’s fixed-width strategy may overlook high-potential paths, illustrating LTM’s flexibility in optimizing resource utilization for improved performance.
\footnote{Appendices \ref{sec:case_study} presents a real illustrative case.}

% Since LTM and beam search generate multiple outputs during a single inference process, we ensure a fair comparison by sampling $n$ outputs for each of the remaining decoding strategies. This approach maintains comparable computational costs by aligning the total number of tokens processed during the model's forward pass. Specifically, for $p^*$ and $k^*$, we test various configurations, calculate the number of tokens generated for each setting, and selecte configurations that matched the token count of other baselines.

% We report the $acc@1$ and $pass@1$ metrics for the outputs with the lowest PPL under comparable computational costs. 

% Nonetheless, it remains prevalent in the current landscape of LLMs owing to the diversity it offers in its outputs.
% Therefore, to enhance the output diversity of LTM while maintaining effectiveness, it is crucial to explore a balance between these two factors in future work.

\section{Related Work}

% In this section, we introduce some related work, including self-correction and decoding methods. 

\subsection{Self-Correction Methods}

% Numerous studies have advanced self-correction methods in the LLMs area. \citet{madaan2024self} proposed a three-stage framework that enhances model outputs by integrating feedback from previous iterations. \citet{li2024confidence} designed prompts that guide the model in assessing its confidence and deciding whether to generate a revised response. \citet{huanglarge} further explored the use of critic prompts to evaluate the self-correction capabilities of LLMs. Additionally, task-specific prompts have been developed for translation tasks to facilitate iterative refinement \citep{chen2023iterative}, and methods like Self-Debug have enabled models to generate feedback based on their own code and execution results \citep{chen2023teaching}. External tools, including search engines and code interpreters, have also been integrated to support error correction \citep{gou2023critic}.

% However, existing self-correction methods for LLMs rely heavily on prompts but neglect real-time user feedback. This can cause redundant refinements and degrade model performance.Unlike previous work, our prompt-free approach integrates useful user feedback to enhance efficiency and performance.

Numerous studies have advanced self-correction methods in the domain of LLMs. For instance, \cite{madaan2024self} proposed a three-stage framework that enhances LLM outputs by integrating feedback from previous iterations. \cite{li2024confidence} designed prompts to guide the LLM in assessing its confidence and deciding whether to generate a revised response. \cite{huanglarge} further explored the use of critic prompts to evaluate the self-correction capabilities of LLMs. Additionally, task-specific prompts have been developed for translation tasks to facilitate iterative refinement \citep{chen2023iterative}.
% Current LLM self-correction methods depend solely on prompts, ignoring user feedback. Our approach eliminates additional prompts and leverages user feedback to boost both efficiency and results.
However, existing self-correction methods for LLMs rely on prompts while neglecting real-time user feedback. This can lead to redundant refinements and degrade LLM performance. Unlike previous work, our prompt-free approach incorporates useful user feedback to enhance efficiency and performance.

\subsection{Decoding Strategies}

Current decoding strategies in LLMs primarily use next-token prediction mechanisms such as top-$k$ sampling \citep{fan2018hierarchical,holtzman2018learning} and nucleus sampling (top-$p$ sampling) \citep{holtzman2019curious}. In top-$k$ sampling, the LLM selects the next token from the top-$k$ most probable tokens, while in nucleus sampling, it samples from the smallest set of tokens whose cumulative probability exceeds a threshold $p$. \cite{basu2020mirostat} proposed a modified version of top-$k$ sampling that incorporates a feedback mechanism to control the perplexity of the generated text. \cite{zhu2024improving} introduced adaptive decoding, a variant of top-$k$ sampling that dynamically adjusts the size of $k$ based on the information entropy of the token probability distribution. Other recent approaches boost LLM performance by exploring multiple decoding paths. For instance, \cite{wang2024chain} combines sampling methods to select optimal outputs, while \cite{zhu2024deductive} scores reasoning steps to determine path expansion. However, these methods are constrained by fixed templates and limited reasoning steps. Our LTM decoding overcomes these limitations through flexible, template-free reasoning for more adaptable human-AI interactions.

% In an effort to improve LLM performance, several approaches have been developed that explore multiple inference paths. For example, \citep{wang2024chain} improved reasoning by combining top-$k$ sampling and greedy sampling, selecting the longest output from the generated candidates. Likewise, \citet{zhu2024deductive} enhanced reasoning by decomposing model outputs into discrete steps and assigning deductive scores to determine how many branches should be processed. However, these approaches are often limited to fixed output templates and constrained reasoning steps, reducing their flexibility and practical applicability. Our proposed LTM decoding addresses these limitations by focusing on long-term reasoning paths and eliminating reliance on predefined templates, enhancing adaptability and flexibility in human-LLM interactions.

\section{Conclusion}
This work introduces the FTR self-correction framework, which significantly enhances the performance of LLMs by leveraging user feedback as a guiding signal. Specifically, when user feedback indicates dissatisfaction with the LLMs' output, the framework employs LTM—an advanced decoding strategy—to refine the response. By integrating LTM with feedback-triggered regeneration, the framework notably improves the overall quality of the LLMs' responses. 
% Unlike traditional prompt-reliant methods using LLM self-assessment, FTR's adaptive design better supports real-world human-LLM interactions through intuitive feedback.
Unlike existing methods that rely heavily on prompts and the LLMs' internal assessment capabilities, the FTR framework is more flexible and adaptive. This makes it particularly well-suited for real-world human-AI interaction scenarios where intuitive feedback is readily available. 

% We propose two directions for future work: (1) Although LTM demonstrates superior performance, it lacks diversity in its outputs. Exploring methods to enhance the diversity of LTM would be a valuable area of research; (2) Recently, reinforcement learning (RL) has gained significant importance in the LLM field. One key direction is improving response generation for possible candidate solution. It would be interesting to investigate whether LTM can aid in generating data that benefits RL.

\section*{Limitations}

% Despite its advantages, LTM is prone to higher levels of repetition and redundancy in text generation. Potential future directions include detecting and eliminating redundant outputs during the decoding process and integrating sampling mechanisms into the algorithm to improve output diversity. These enhancements could further reduce computational load and latency, thereby improving real-time user interaction.

% In addition, our current experiments focus on LLMs with fewer than 13 billion parameters, as these models provide a cost-effective testbed for method validation. While prior research has demonstrated the efficacy of prompt-based methods in larger-scale models (e.g., OpenAI GPT series) due to their enhanced instruction-following and reasoning capacities \cite{li2024confidence}, our controlled experiments with smaller models establish important baselines. Future work will systematically evaluate the scalability of our approach across model sizes, particularly in larger LLM architectures.

Despite its advantages, LTM’s multipath decoding mechanism exhibits a tendency to produce repetitive or redundant outputs in text generation tasks. Future research may focus on developing mechanisms to detect and mitigate such redundancy during decoding—such as dynamic path pruning based on semantic similarity—while incorporating advanced sampling techniques (e.g., diverse beam search) to enhance output diversity. These enhancements could not only reduce computational overhead and inference latency but also improve real-time interaction quality in human-AI dialogue systems.
Meanwhile, our current experimental scope is limited to LLMs with fewer than 13 billion parameters, which serve as a cost-efficient platform for method validation. While prior studies \cite{li2024confidence} have shown that prompt-based approaches excel in larger models (e.g., GPT series) due to their superior instruction-following capabilities, our controlled experiments on smaller architectures provide essential baselines for low-resource scenarios. Future work will systematically evaluate LTM’s scalability across LLM model scales to characterize its impacts on computational efficiency, accuracy, and latency in diverse computational environments.

%Bibliography
\bibliographystyle{unsrt}  
\bibliography{references}  

\appendix
\section{Appendices}

\subsection{Case Study}
\label{sec:case_study}
\begin{figure}[h]
\centering
\includegraphics[width=0.6\textwidth]{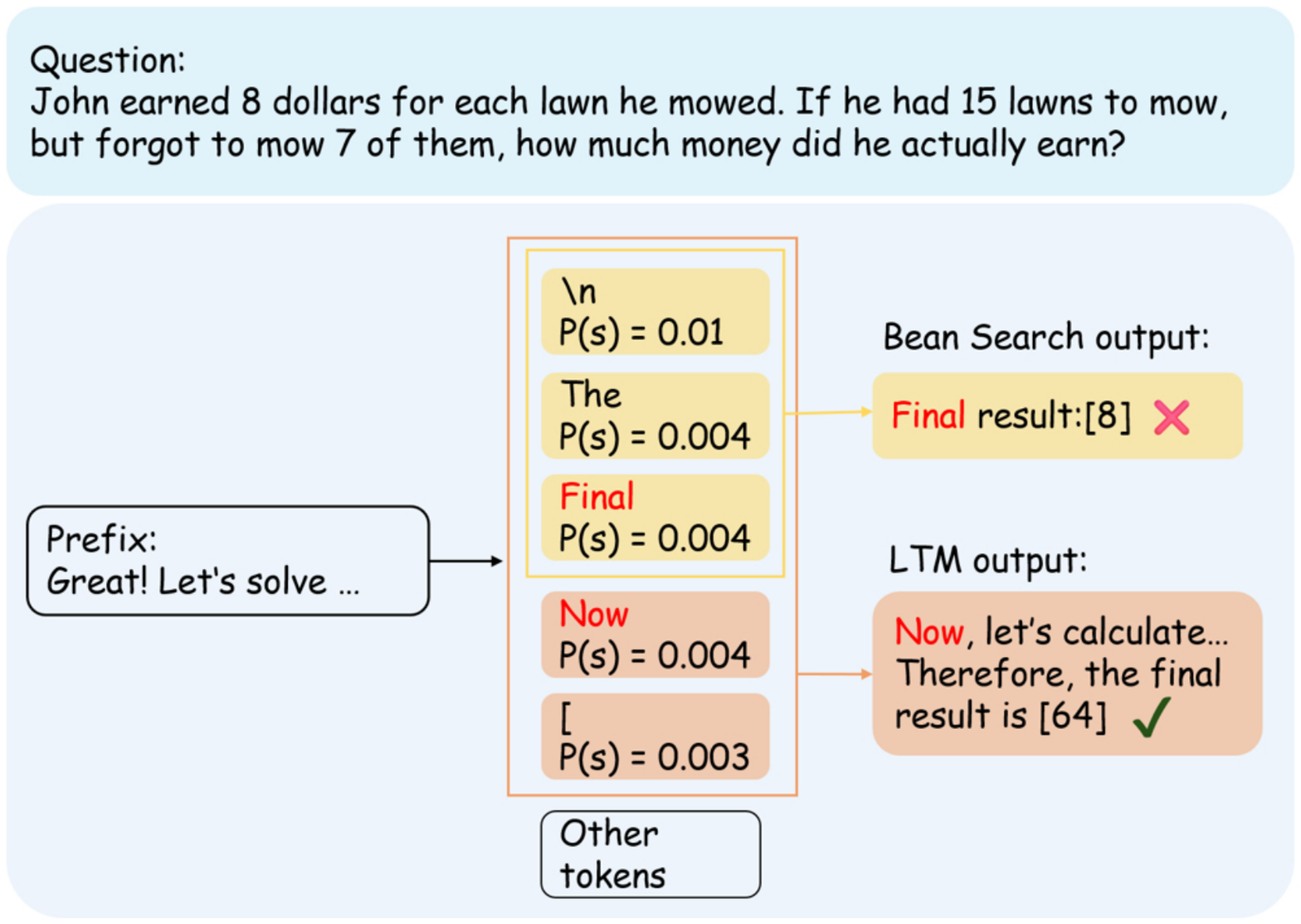} % Reduce the figure size so that it is slightly narrower than the column.
\caption{Comparison of decoding processes between Beam Search (fixed beam width=3) and LTM on a MultiArith dataset question. The yellow rectangle denotes Beam Search's candidate set, while the orange rectangle represents LTM's candidate set.}
\label{case}
\end{figure}

In this section, we present a case study comparing LTM with Beam Search to demonstrate LTM's distinctive advantages. As illustrated in Figure~\ref{case}, we analyze the reasoning processes of both methods on a mathematical problem from the MultiArith dataset. For controlled comparison, Beam Search uses a fixed beam width of 3, while LTM dynamically adjusts its beam width with an average value of 3 across decoding steps.

The key observation reveals that at a critical decoding step with a relatively flat candidate distribution, Beam Search fails to retain all promising candidates due to its fixed beam width strategy, ultimately leading to an incorrect solution. In contrast, LTM dynamically adjusts its computational resources at this step and increases the beam width to 5, successfully preserving all potentially correct candidates and consequently arriving at the accurate final answer. This case clearly demonstrates LTM's adaptive advantage over static beam search methods in handling complicated decoding scenarios.

\subsection{Inference Time Analysis}
% Although LTM incurs computational overhead versus single-beam decoding, this cost is limited to negative feedback cases in our FTR framework. Unlike standard self-correction methods that always regenerate outputs, our approach selectively triggers regeneration, yielding net performance gains.

% The Critic and IoE Prompt methods require $2Nt$ computation time, where $N$ represents the number of samples in the dataset and $t$ denotes the average LLM response time per question. This quadratic complexity arises because these methods perform two complete generation passes for each sample. In contrast, our FTR framework requires $Nt(1+pn)$ computation time for its two-stage regeneration process, where $p$ is the percentage of samples requiring correction and $n$ is the average LTM beam width. This approach becomes more efficient when $pn<1$. 
% Through parallel LTM decoding, we achieve additional inference time reductions. For instance, with LTM decoding at an average beam width of 3, the processing speed is approximately 1.85 times of single-beam decoding. This means that when $p < 0.54$, FTR's total computation time becomes shorter than that of both Critic and IoE Prompt-based methods. Detailed timing comparisons across different models and datasets are presented in Table \ref{speed}.

Although LTM decoding introduces computational overhead compared to single-beam decoding, FTR’s design mitigates this cost by triggering regeneration only for samples with negative feedback, unlike standard self-correction methods that perform mandatory double-pass generation for all inputs and thus lead to net efficiency gains in low error rate scenarios. Baseline methods such as Critic and IoE Prompt require \(2 \times N \times t\) computation time (with \(N\) as the number of samples and \(t\) as the average inference time per sample), a linear complexity arising from re-generating responses for all inputs. In contrast, FTR’s total computation time is \(N \times t \times (1 + p \times n)\), where \(p\) is the proportion of samples requiring regeneration and \(n\) is the average LTM beam width, with efficiency emerging when \(p \times n < 1\). 
It is worth noting that the impact of beam size on performance can be reduced through parallel computing.
For example, with parallel LTM decoding at an average beam width \(n=3\), the per-sample regeneration cost is \(1.85\times\) single-beam decoding time, leading to the efficiency condition \(p < \frac{1}{1.85} \approx 0.54\). 
Detailed test results across backend LLMs and datasets are presented in Table \ref{speed}, confirming that FTR’s adaptive regeneration strategy balances reasoning depth with computational efficiency for real-world deployments.
\footnote{The experimental evaluations were performed on a computing system featuring an NVIDIA RTX 6000 Ada Generation GPU and Intel Xeon Platinum 8358P CPUs with 32 physical cores (64 threads) running at 2.60GHz base frequency and 48MB L3 cache.
}

\begin{table}[h]
  \centering
    \begin{tabular}{cccc}
    \toprule
    Method & \textbf{GSM8K} & \textbf{MultiArith} & \textbf{HumanEval} \\
    \midrule
    \rowcolor{gray!20}
    \multicolumn{4}{c}{\# Llama2-7B \#} \\
    Initial Input & $1\times$ & $1\times$ & $1\times$  \\
    + Critic/IoE Prompt & $2\times$ & $2\times$ & $2\times$ \\
    + FTR (Ours)& $ 2.47\times$ & $ 2.15\times$ & $ 3.87\times$ \\
    \rowcolor{gray!20}
    \multicolumn{4}{c}{\# Llama2-13B \#} \\
    Initial Input & $1\times$ & $1\times$ & $1\times$  \\
    + Critic/IoE Prompt & $2\times$ & $2\times$ & $2\times$ \\
    + FTR (Ours)& $ 2.48\times$ & $ 1.64\times$ & $ 2.98\times$ \\
    % DeepSeek-1.5B& 0.569 & 0.850 & -  \\
    % DeepSeek-7B&  & & -  \\
      \rowcolor{gray!20}
    \multicolumn{4}{c}{\# Llama3-1B \#} \\
    Initial Input & $1\times$ & $1\times$ & $1\times$  \\
    + Critic/IoE Prompt & $2\times$ & $2\times$ & $2\times$ \\
    + FTR (Ours)& $ 2.40\times$ & $ 2.49\times$ & $ 2.32\times$ \\
    % DeepSeek-1.5B& 0.604 & 0.883 & -  \\
    % DeepSeek-7B&  & & -  \\
          \rowcolor{gray!20}
    \multicolumn{4}{c}{\# Llama3-3B\#} \\
    Initial Input & $1\times$ & $1\times$ & $1\times$  \\
    + Critic/IoE Prompt & $2\times$ & $2\times$ & $2\times$ \\
    + FTR (Ours)& $ 1.42\times$ & $ 1.12\times$ & $ 2.64\times$ \\
              \rowcolor{gray!20}
    \multicolumn{4}{c}{\# Qwen-1.5B\#} \\
    Initial Input & $1\times$ & $1\times$ & $1\times$  \\
    + Critic/IoE Prompt & $2\times$ & $2\times$ & $2\times$ \\
    + FTR (Ours)& $2.07 \times$ & $ 1.81\times$ & $ 2.89\times$ \\
              \rowcolor{gray!20}
    \multicolumn{4}{c}{\# Qwen-3B\#} \\
    Initial Input & $1\times$ & $1\times$ & $1\times$  \\
    + Critic/IoE Prompt & $2\times$ & $2\times$ & $2\times$ \\
    + FTR (Ours)& $ 1.30\times$ & $1.02\times$ & $ 2.03\times$ \\
    % DeepSeek-1.5B& \textbf{0.824} & \textbf{1.000} & -  \\
    % DeepSeek-7B&  & & -  \\
    \bottomrule
    \end{tabular}%
    \caption{Inference time comparison of different self-correction methods across different datasets and LLMs.}
  \label{speed}%
\end{table}%

\subsection{Additional Experiments Details}

\begin{table}[h]
\centering
\begin{tabular}{ccccc}
  \toprule
  LLM Model &  & \textbf{GSM8K} & \textbf{MultiArith} & \textbf{HumanEval}  \\
  \midrule
  \multirow{2}{*}{Llama2-7B}& 
    FN & 34.93 & 13.83 & 5.88\\
    & FP & 8.40 & 30.23 & 1.36 \\
    \midrule
    \multirow{2}{*}{Llama2-13B}& 
    FN & 22.31 & 22.03 & 11.76\\
    & FP & 9.57 &40.32  & 1.54 \\
    \midrule
    \multirow{2}{*}{Llama3-1B}& 
    FN & 16.41 & 2.74 & 6.38\\
    & FP & 25.71 & 60.75 & 5.13 \\
    \midrule
    \multirow{2}{*}{Llama3-3B}& 
    FN & 13.03 & 1.73 & 15.00\\
    & FP & 15.77 & 28.57 & 17.86 \\
    \midrule
    \multirow{2}{*}{Qwen-1.5B}& 
    FN & 13.54 & 4.10 & 20.90\\
    & FP & 35.70 & 82.76 & 20.62 \\
    \midrule
    \multirow{2}{*}{Qwen-3B}& 
    FN & 9.51 & 0.56 & 8.11\\
    & FP & 18.14 & 0.00 & 20.75 \\
    
  \bottomrule
\end{tabular}

\caption{False Negative (FN) and False Positive (FP) rates of GPT-4o on different datasets (\%).}
\label{falsenegative}
\end{table}

\begin{table}[h]
  \centering
    \begin{tabular}{cccc}
    \toprule
    Method & \textbf{GSM8K} & \textbf{MultiArith} & \textbf{HumanEval} \\
    \midrule
    \rowcolor{gray!20}
    \multicolumn{4}{c}{\# Llama2-7B \#} \\
    Initial Input & 0.206 & 0.539 & 0.104  \\
    + Prompt & 0.243 & 0.694 & 0.140 \\
    + Indicator (Ours) & 0.328 & 0.810 & 0.146  \\
    \rowcolor{gray!20}
    \multicolumn{4}{c}{\# Llama2-13B \#} \\
    Initial Input & 0.303 & 0.656 & 0.207  \\
    + Prompt & 0.359 & 0.806 & 0.220 \\
    + Indicator (Ours) & 0.455 & 0.872 & 0.243  \\
    % DeepSeek-1.5B& 0.569 & 0.850 & -  \\
    % DeepSeek-7B&  & & -  \\
      \rowcolor{gray!20}
    \multicolumn{4}{c}{\# Llama3-1B \#} \\
    Initial Input & 0.245 & 0.406 & 0.287  \\
    + Prompt & 0.301 & 0.506 & 0.317 \\
    + Indicator (Ours)& 0.374 & 0.556 & 0.384  \\
    % DeepSeek-1.5B& 0.604 & 0.883 & -  \\
    % DeepSeek-7B&  & & -  \\
          \rowcolor{gray!20}
    \multicolumn{4}{c}{\# Llama3-3B\#} \\
    Initial Input & 0.774 & 0.961 & 0.488  \\
    + Prompt & 0.822 & 0.972 & 0.534 \\
    + Indicator (Ours) & 0.859 & 0.983 & 0.567  \\
              \rowcolor{gray!20}
    \multicolumn{4}{c}{\# Qwen-1.5B\#} \\
        Initial Input & 0.422 & 0.678 & 0.409  \\
    + Prompt & 0.512 & 0.750 & 0.476 \\
    + Indicator (Ours) & 0.630 & 0.900 & 0.512  \\
              \rowcolor{gray!20}
    \multicolumn{4}{c}{\# Qwen-3B\#} \\
    Initial Input & 0.837 & 0.994 & 0.677  \\
    + Prompt & 0.867 & 0.994 & 0.732 \\
    + Indicator (Ours) & 0.892 & 1.000 & 0.768  \\
    % DeepSeek-1.5B& \textbf{0.824} & \textbf{1.000} & -  \\
    % DeepSeek-7B&  & & -  \\
    \bottomrule
    \end{tabular}%
    \caption{Detailed model performance comparison under different user feedback utilization approaches.}
  \label{feedbackuse}%
\end{table}%

\begin{table*}[h]
\centering
\begin{tabular}{cccc}
\toprule
   & GSM8K & MultiArith & HumanEval \\
\midrule
\rowcolor{gray!20}
    \multicolumn{4}{c}{\# Llama2-7B \#} \\
    Beam Search & $n =3$ & $n =5$ & $n =7$ \\
 Combined Sampling &$n =3$ & $n =5$& $n =6$\\
 Adaptive Decoding &$n =3$ &$n =5$ & $n =7$ \\
 LTM Decoding & $p^* =0.8 , k^* =7 $ &$p^* =0.9 , k^* = 7$ & $p^* =0.85 , k^* = 7$ \\
\rowcolor{gray!20}
    \multicolumn{4}{c}{\# Llama2-13B \#} \\
    Beam Search & $n =4$ & $n =3$ & $n =5$ \\
 Combined Sampling &$n =4$ & $n =3$& $n =5$\\
 Adaptive Decoding &$n =4$ &$n =3$ & $n =5$ \\
 LTM Decoding & $p^* =0.9 , k^* =7 $ &$p^* = 0.85, k^* =7 $ & $p^* =0.8 , k^* = 7$ \\
\rowcolor{gray!20}
    \multicolumn{4}{c}{\# Llama3-1B \#} \\
    Beam Search & $n =3$ & $n =5$ & $n =3$ \\
 Combined Sampling &$n =3$ & $n =5$& $n =3$\\
 Adaptive Decoding &$n =3$ &$n =5$ & $n =3$ \\
 LTM Decoding & $p^* =0.8 , k^* =7 $ &$p^* =0.9 , k^* = 7$ & $p^* = 0.8, k^* =7 $ \\
\rowcolor{gray!20}
    \multicolumn{4}{c}{\# Llama3-3B \#} \\
Beam Search & $n =3$ & $n =7$ & $n =5$ \\
 Combined Sampling &$n =3$ & $n =6$& $n =7$\\
 Adaptive Decoding &$n =3$ &$n =6$ & $n =7$ \\
 LTM Decoding & $p^* = 0.8, k^* = 7$ &$p^* =0.95 , k^* = 7$ & $p^* =0.9 , k^* = 7$ \\
\rowcolor{gray!20}
    \multicolumn{4}{c}{\# Qwen-1.5B \#} \\
   Beam Search & $n =3$ & $n =5$ & $n =7$ \\
 Combined Sampling &$n =3$ & $n =5$& $n =7$\\
 Adaptive Decoding &$n =3$ &$n =5$ & $n =7$ \\
 LTM Decoding & $p^* =0.8 , k^* = 7$ &$p^* =0.9 , k^* =7 $ & $p^* = 0.9, k^* = 8$ \\
\rowcolor{gray!20}
    \multicolumn{4}{c}{\# Qwen-3B \#} \\
    Beam Search & $n =3$ & $n =5$ & $n =7$ \\
 Combined Sampling &$n =3$ & $n =5$& $n =7$\\
 Adaptive Decoding &$n =3$ &$n =5$ & $n =7$ \\
 LTM Decoding & $p^* =0.85 , k^* =7 $ &$p^* = 0.85, k^* = 7$ & $p^* =0.9 , k^* = 7$ \\
\bottomrule
\end{tabular}
\caption{Hyperparameters of different decoding methods used in Tables~\ref{selfcorrect} and~\ref{ablation}. For beam search, $n$ denotes the number of beams, while for combined sampling and adaptive decoding, $n$ indicates the number of generated answers.}
\label{hyperparameter}
\end{table*}

\end{document}